\pdfoutput=1
\UseRawInputEncoding
\documentclass[10pt,twocolumn,letterpaper]{article}

\usepackage[pagenumbers]{iccv} 
\usepackage{amsmath,amsfonts}
\usepackage{array}
\usepackage{textcomp}
\usepackage{stfloats}
\usepackage{url}
\usepackage{verbatim}
\usepackage{multirow}
\usepackage[linesnumbered,ruled,vlined,ruled]{algorithm2e}
\usepackage{booktabs}
\usepackage{color}
\usepackage{longtable}
\usepackage{graphicx}
\usepackage{bbding}
\usepackage{caption,subcaption}
\usepackage{threeparttable}
\usepackage{tabularx}
\usepackage{arydshln} 
\newcolumntype{R}[1]{>{\raggedleft\arraybackslash}p{#1}}
\newcolumntype{L}[1]{>{\raggedright\arraybackslash}p{#1}}
\newcolumntype{C}[1]{>{\centering\arraybackslash}p{#1}}
\newcolumntype{Y}{>{\raggedright\arraybackslash}X} 
\newcolumntype{Z}{>{\raggedleft\arraybackslash}X}
\newcolumntype{A}{>{\centering\arraybackslash}X}
\usepackage{tikz}
\usetikzlibrary{shapes, arrows, positioning}
\tikzstyle{startstop} = [rectangle, rounded corners, minimum width=3cm, minimum height=1cm, text centered, draw=black, fill=red!30]
\tikzstyle{io} = [trapezium, trapezium left angle=70, trapezium right angle=110, minimum width=3cm, minimum height=1cm, text centered, draw=black, fill=blue!30]
\tikzstyle{process} = [rectangle, minimum width=3cm, minimum height=1cm, text centered, draw=black, fill=orange!30]
\tikzstyle{arrow} = [thick,->,>=stealth]
\usepackage{booktabs} 
\usepackage{siunitx} 
\usepackage{makecell}
\usepackage{algpseudocode}
\usepackage{amsmath}
\usepackage{graphicx}
\usepackage{subcaption}
\usepackage{caption}
\usepackage{adjustbox}
\definecolor{iccvblue}{rgb}{0.21,0.49,0.74}
\usepackage[pagebackref,breaklinks,colorlinks,allcolors=iccvblue]{hyperref}
\def\paperID{15740} 
\def\confName{ICCV}
\def\confYear{2025}

\title{SemanticFlow: A Self-Supervised Framework for Joint Scene Flow Prediction and Instance Segmentation in Dynamic Environments}

\author{Yinqi Chen\\
\\
\\
\and
Meiying Zhang\\
\\
\\
\and
Qi Hao\\
\\
\\
\and
Guang Zhou\\
\\
\\
}

\begin{document}
\maketitle
\begin{abstract}
Accurate perception of dynamic traffic scenes is crucial for high-level autonomous driving systems, requiring robust object motion estimation and instance segmentation. However, traditional methods often treat them as separate tasks, leading to suboptimal performance, spatio-temporal inconsistencies, and inefficiency in complex scenarios due to the absence of information sharing. This paper proposes a multi-task SemanticFlow framework to simultaneously predict scene flow and instance segmentation of full-resolution point clouds. The novelty of this work is threefold: 
1) developing a coarse-to-fine prediction based multi-task scheme, 
where an initial coarse segmentation of static backgrounds and dynamic objects is used to provide contextual information 
for refining motion and semantic information through a shared feature processing module; 
2) developing a set of loss functions to enhance the performance of scene flow estimation and instance segmentation, 
while can help ensure spatial and temporal consistency of both static and dynamic objects within traffic scenes; 
3) developing a self-supervised learning scheme, 
which utilizes coarse segmentation to detect rigid objects and compute their transformation matrices between sequential frames, 
enabling the generation of self-supervised labels.
The proposed framework is validated on the Argoverse and Waymo datasets, demonstrating superior performance in instance segmentation accuracy, scene flow estimation, and computational efficiency, establishing a new benchmark for self-supervised methods in dynamic scene understanding.
\end{abstract}    
\section{Introduction}
The rapid development of autonomous driving and intelligent transportation systems necessitates advanced dynamic scene understanding, with scene flow prediction \cite{zhai2021optical} playing a central role in capturing 3D motion and object interactions. While significant progress has been made in both scene flow estimation and point cloud instance segmentation, the treatment of these tasks in isolation within dynamic environments presents significant challenges. Achieving effective scene understanding in real-world applications requires not only accurate motion estimation but also a comprehensive interpretation of object boundaries. Yet, current approaches face three interrelated challenges:
\begin{figure*}[t]
  \centering
   \includegraphics[width=0.9\linewidth]{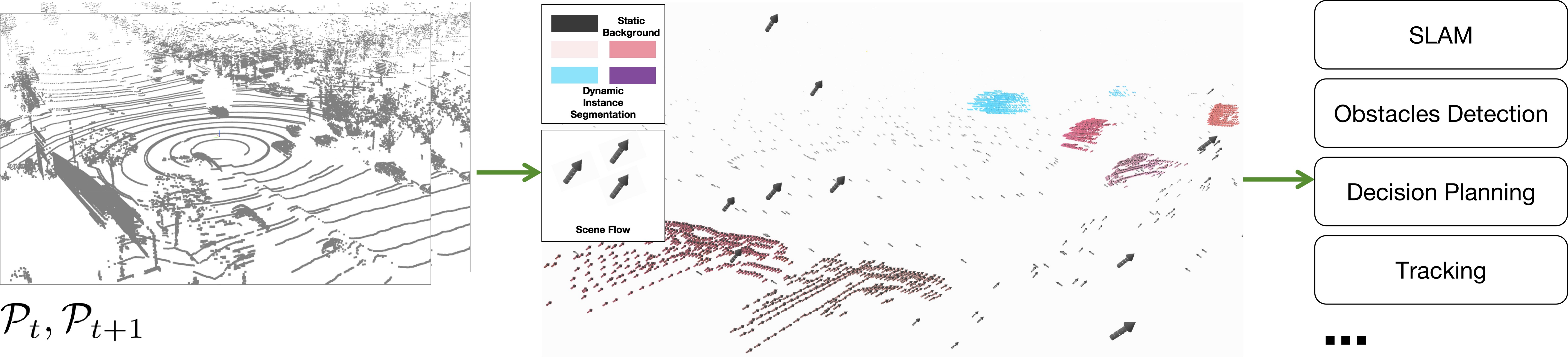}
   \caption{An illustration of SemanticFlow, which estimates scene flow and performs instance segmentation from two consecutive point clouds, enabling applications in SLAM\cite{bahraini2018slam}, obstacle detection, decision planning, tracking and etc..}
   \label{fig:introduction}
\end{figure*}

\begin{itemize}
    \item \textbf{Lack of Mutual Enhancement between Scene Flow Estimation and Instance Segmentation:}  
    Recent studies \cite{gojcic2021weakly, NEURIPS2022_c6e38569} demonstrate that scene flow prediction and instance segmentation are mutually beneficial. 
    However, most methods \cite{kabir2025terrain,jund2021scalablesceneflowpoint,zhang2024deflow,li2023fast,vedder2023zeroflow,lin2024icp,zhang2024seflow,Shi_2019_CVPR,huang2021multibodysync} treat these tasks independently, limiting efficiency.
    Some methods \cite{Baur2021ICCV,zhai2025dmrflow} combine scene flow estimation with foreground-background segmentation without instance-level precision and cannot distinguish whether a given label is foreground or background, and require extra post-processing. 
    Although SSF-PAN \cite{chen2025ssfpansemanticsceneflowbased} integrates the two tasks, it processes them in separate stages without high efficiency. 
    This highlights the need for more integrated approaches to better exploit the synergies between scene flow estimation and instance segmentation. 
    
    \item \textbf{Imbalance between Segmentation Precision and Point Cloud Sampling:}  
    Fine-grained instance segmentation of objects can lead to high computational complexity, 
    while many scene flow estimation methods downsample the point cloud to mitigate computational costs, 
    such as the architecture of PointNet \cite{qi2017pointnet++,wang2023active}. 
    However, this downsampling can severely compromise the precision of motion estimation for individual objects, 
    as the semantic segmentation becomes sparse and less accurate. 
    For instance, Jund et al.  \cite{jund2021scalablesceneflowpoint} showed that 50\% downsampling increases L2 error by up to 159\%. 
    This highlights the challenge of balancing segmentation precision and motion estimation accuracy, 
    which requires a proper loss function design which can effectively integrate both segmentation and scene flow prediction, 
    achieving better balance between efficiency and precision.   
    \item \textbf{Short of Scene Flow and Instance Segmentation Labels:}
    Usually, scene flow estimation relies on semantic segmentation labels to improve motion prediction \cite{gojcic2021weakly, li2022rigidflow,moosmann2010motion}, while point cloud instance segmentation requires motion information to enhance segmentation accuracy \cite{NEURIPS2022_c6e38569, lentsch2024union}. However, both types of data are difficult to annotate, leading to a short of required labeled scene flow datasets.

    \end{itemize}

As shown in \cref{fig:introduction}, this paper proposes a 
framework that jointly optimizes scene flow and instance segmentation through innovative loss functions.
The contributions of this paper are as follows:
\begin{enumerate}
    \item Developing a multi-task framework which utilizes motion features to distinguish the background from dynamic objects, 
    directly supporting downstream tasks such as SLAM\cite{bahraini2018slam}, navigation, and tracking.
    It utilizes a coarse-to-fine strategy, where an initial coarse segmentation guides scene flow estimation, 
    and the precise motion and segmentation information are obtained through a shared refinement process.

    \item Developing a set of loss functions based on the object-level consistency and dynamic region contrast, 
    which enforce motion coherence within segmented instances, 
    bridging the gap between semantic understanding and motion prediction.
    
    \item Developing a self-supervised framework which generates pseudo-labels for both scene flow and instance segmentation
    by leveraging implicit motion cues and temporal consistency, 
    eliminating the need for manual annotations and enabling training without additional labels.
    
    \item The proposed SemanticFlow achieves state-of-the-art performance in self-supervised scene flow estimation on the Argoverse 2 leader board, outperforming most supervised methods in key metrics.\footnote{\url{https://eval.ai/web/challenges/challenge-page/2210/leaderboard/5463}}.
\end{enumerate}

\section{Related Work}
\subsection{Multi-Task Learning for Scene Flow Estimation and Instance Segmentation}
Multi-task learning (MTL) \cite{crawshaw2020multitasklearningdeepneural} has been widely used in autonomous driving and 3D perception to improve efficiency and generalization through shared representations. 
SLIM \cite{Baur2021ICCV} incorporates motion segmentation into self-supervised scene flow by exploiting the discrepancy between rigid ego-motion and flow predictions, but it only classifies foreground and background, 
limiting its ability to distinguish dynamic objects. 
SFEMOS \cite{chen2024joint} jointly addresses scene flow estimation and moving object segmentation by partitioning
the scene into static and moving regions, but it requires labeled MOS data, 
limiting its real-world applicability. 
SemanticFlow establishes bidirectional dependencies between tasks, 
enabling task-aware feature learning that enhances motion prediction and segmentation robustness in complex traffic environments. 

\subsection{Interdependent Loss Functions for Efficient Scene Flow and Semantic Segmentation}

The balance between precision and efficiency in scene understanding remains a core challenge. 
Multi-stage methods decompose complex tasks to improve precision. 
In scene flow estimation, piecewise rigid transformation methods, 
such as those by Gojcic et al. ~\cite{gojcic2021weakly} and Li et al.~\cite{li2022rigidflow}, iteratively estimate rigid motions and optimize segmentation. 
However, these methods struggle with non-rigid objects and convergence speed in real-time scenarios. 
Coarse-to-fine strategies like PWC-Net~\cite{Sun2018PWC-Net} improve efficiency through progressive optimization. 
Inspired by these methods, the proposed SemanticFlow adopts a multi-stage strategy with interdependent loss functions: 
it first performs coarse segmentation, 
followed by scene flow recovery on full-resolution point clouds, and then refines segmentation accuracy based on scene flow, effectively balancing computational efficiency and motion precision. 

\subsection{Self-Supervised Scene Flow and Instance Segmentation}

Many methods, like PointPWC-Net \cite{wu2020pointpwc}, use hierarchical cost volumes for 3D motion prediction, while FLOT \cite{puy2020flot} applies optimal transport for point matching, but both are supervised. Self-supervised learning has since emerged as a solution to alleviate the annotation burden. Iterative methods, such as FlowStep3D \cite{kittenplon2021flowstep3d} and PV-RAFT \cite{wei2021pv}, refine predictions via GRUs\cite{dey2017gate} but lack segmentation integration. Weakly supervised approaches{ \cite{gojcic2021weakly,zhang2024seflow,10906337}, rely on rigid motion constraints or supervised odometry, struggling with non-rigid dynamics and fine-grained segmentation. Clustering-based methods like DBSCAN \cite{ester1996density} and WardLinkage \cite{ward1963hierarchical} group points based on geometric and motion similarities but fall short in dynamic object segmentation. Deep learning methods \cite{huang2021multibodysync, yi2018deep} are often limited to binary foreground-background separation or require supervision, while methods like Thomas et al. \cite{thomas2021self} and OGC \cite{NEURIPS2022_c6e38569} face challenges in accurately segmenting dynamic objects and isolating the static background. In contrast, the proposed SemanticFlow leverages a shared encoder and task-specific decoders, and eliminates the need for external labels and improves object-level motion reasoning, 
achieving superior accuracy and scalability over previous methods.



\begin{figure*}[t]
  \centering
   \includegraphics[width=1.0\linewidth]{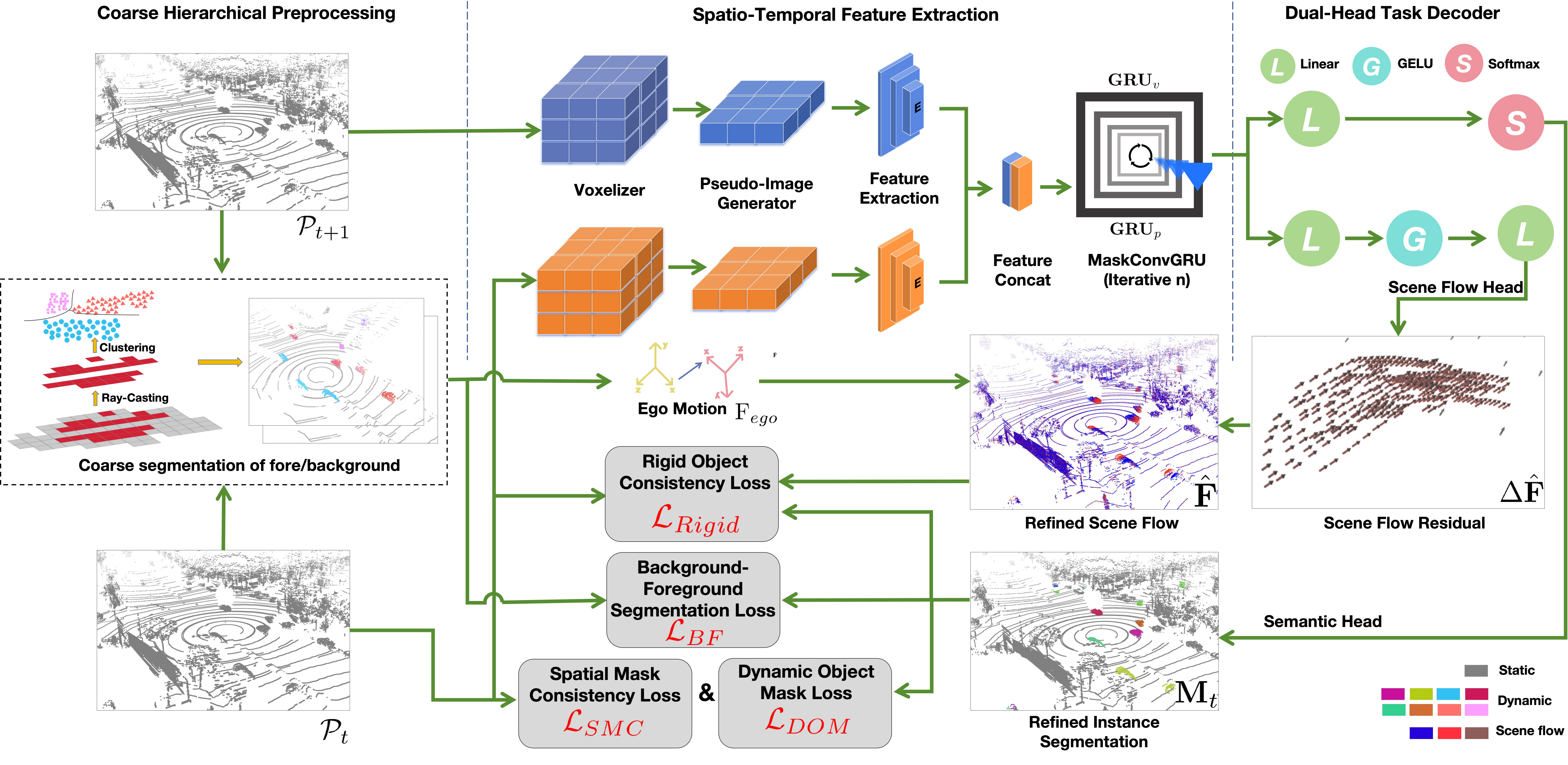}
   \caption{ An illustration of the SemanticFlow system diagram.}
   \label{fig:system}
\end{figure*}
\section{System setup and problem statement}
\subsection{System Setup}
Our system uses a self-supervised multi-task learning framework to simultaneously perform scene flow prediction and instance segmentation, as illustrated in \cref{fig:system}. 
\subsubsection{Input and Output}
\begin{itemize}
\item{Input}
\begin{itemize}
    \item \(\mathcal{P}_t = \{p_{t,i}\}_{i=1}^N\): The point cloud data at frame \(t\), containing \(N\) points \(p_{t,i}\), where \(i\) denotes the point index.
    \item \(\mathcal{P}_{t+1} = \{p_{t+1,i}\}_{i=1}^N\): The point cloud at frame \(t+1\).
\end{itemize}

\item{Output}
\begin{itemize}
    \item \textbf{Scene Flow Residual \(\Delta \hat{\mathbf{F}}\)}:
     It is predicted by the network, representing local motion in the dynamic environment under the assumption that the ego vehicle is stationary. The final predicted scene flow \(\hat{\mathbf{F}}\) is decomposed into two parts:
     \begin{equation}
    \hat{\mathbf{F}} = \mathbf{F}_\text{ego} + \Delta \hat{\mathbf{F}}
    \label{eq:sceneflow}
    \end{equation}
    where\(\mathbf{F}_\text{ego}\) is estimated by computing the ego vehicle’s motion from the background point cloud after coarse segmentation of two consecutive frames.
    
    \item \textbf{Refined Instance Segmentation \(\mathbf{M}_t \in \mathbb{R}^{N \times C}\)}: It represents the output of the mask classification logits, where \(C\) is the number of categories. For each point \(i\), the logits vector \(\mathbf{m}_i \in \mathbb{R}^{C}\) contains unnormalized probabilities for each category, expressed as \(\mathbf{m}_i = [m_{i,0}, m_{i,1}, \dots, m_{i,C-1}]\). The matrix \(\mathbf{M}_t\) can be written as \([\mathbf{m}_1, \mathbf{m}_2, \dots, \mathbf{m}_N]^T\).
    The final predicted classification label \(\hat{l}_i\) for each point \(i\) is determined by applying the \(\arg\max\) operation on \(\mathbf{m}_i\):
    \[
    \hat{l}_i = \arg\max(\mathbf{m}_i)
    \]
    
    The classification rules are as follows:
    \[
    \hat{l}_i = 
    \begin{cases} 
    0 & \text{background static point}\\
    1, 2, \dots, C-1 & \text{foreground dynamic point}.
    \end{cases}
    \]
    \end{itemize}
\end{itemize}
\subsubsection{Framework Pipeline}
The system framework consists of the following components:

\begin{itemize}
    \item \textbf{Coarse Hierarchical Preprocessing:} Unsupervised preprocessing first uses ray-casting for coarse separation of static and dynamic points, then applies clustering to segment dynamic points.
    \item \textbf{Spatio-Temporal Feature Extraction:} A unified encoder extracts global geometric features from input point clouds, serving as a shared backbone for both tasks. To refine features, a recurrent GRU-based module iteratively updates voxel-wise and point-wise representations. Specifically, $\text{GRU}_v$ aggregates point features within each voxel, refining voxel-level representations, while $\text{GRU}_p$ updates point features using the refined voxel information. This iterative process enables efficient information propagation across scales, reducing redundant computations while preserving spatial details.
    \item \textbf{Dual-Head Task Decoder:} 
\begin{itemize}
    \item \textit{Scene Flow Decoder:} This module estimates the 3D motion vectors for each point by utilizing both global and local geometric features.
    \item \textit{Instance Segmentation Decoder:} This decoder segments the scene into static and dynamic regions and performs fine-grained segmentation of foreground dynamic object instances.
\end{itemize}
\end{itemize}

\subsubsection{Loss Function Design}
The multi-task loss function incorporates the following components:

\begin{itemize}
    \item \textbf{Scene Flow Loss:} Ensures accurate prediction of motion vectors for both static and dynamic points, including the Chamfer Distance Loss for scene flow alignment.
    \item \textbf{Segmentation Loss:} Guides the model in distinguishing between static and dynamic points, including the Background-Foreground Segmentation Loss (\(\mathcal{L}_{\text{BF}}\)) and Dynamic Object Mask Loss (\(\mathcal{L}_{\text{DOM}}\)).
    \item \textbf{Task Synergy Constraints:} Enforces consistency between scene flow estimation and segmentation, incorporating Rigid Object Consistency Loss (\(\mathcal{L}_{\text{Rigid}}\)) and Spatial Mask Consistency Loss (\(\mathcal{L}_{\text{SMC}}\)).
\end{itemize}

\subsection{Problem Statement}

In the context of autonomous driving and dynamic scene understanding, our approach addresses the following key challenges:
\begin{itemize}
    \item How to leverage unsupervised coarse data preprocessing and pseudo-label generation techniques to effectively split static and dynamic regions, refining the segmentation of dynamic objects without relying on labeled data?
    \item How to construct loss functions that simultaneously predicts scene flow and performs instance segmentation, ensuring both tasks benefit from shared feature extraction while maintaining task-specific optimization?
    \item How to handle discrepancies between coarse and fine data via the self-supervised framework to generate more precise predictions?
\end{itemize}
\section{Proposed Methods}

\begin{algorithm}[t]
\caption{Coarse-to-Fine Scheme}
\label{alg:coarse_to_fine}
\KwIn{Consecutive point clouds $\mathcal{P}_t, \mathcal{P}_{t+1} \in \mathbb{R}^{N \times 3}$}

\textbf{Step 1: Coarse Dynamic Segmentation}  

Compute dynamic probability $\psi(p_{t,i})$ via \cref{eq:dufomap} \;  
Extract dynamic points: $\mathcal{P}_t^d \gets \{p_{t,i} \mid \psi(p_{t,i}) = 1\}$ \;  
Cluster $\mathcal{P}_t^d$ into $\mathcal{C}_k^{(t)}$ via DBSCAN \;  

\textbf{Step 2: Voxel-Based Representation}  

Compute voxel coordinates $\{v\}$ via \cref{eq:voxel} \;  
Initialize point features: $\mathbf{f}_{t,i}^{(0)} \gets \text{MLP}(p_{t,i})$ \;  
Initialize voxel states: $\mathbf{h}_{v,t}^{(0)} \gets \mathbf{0}, \forall v \in \mathcal{V}_t$ \;  

\textbf{Step 3: Feature Refinement with GRU}  

\For{$k = 0$ \textbf{to} $K-1$}{  
    Update voxel states $\mathbf{h}_{v,t}^{(k+1)}$, point features $\mathbf{f}_{t,i}^{(k+1)}$ via \cref{eq:gru_update} \;  
}
\end{algorithm}
\subsection{Coarse-to-Fine Instance Segmentation}
\label{subsec:pseudo_label}

\subsubsection{Coarse Segmentation of Background and Foreground}
Following Step 1 in \cref{alg:coarse_to_fine}, given consecutive point cloud frames $\mathcal{P}_t$ and $\mathcal{P}_{t+1}$, the dynamic probability is computed as:
\begin{equation}
\psi(p_{t,i}) = \begin{cases}
1 & \text{if } \frac{\|p_{t+1,i} - \mathcal{R}(p_{t,i})\|}{\|p_{t,i}\|} > \tau_r \\
0 & \text{otherwise}
\end{cases}
\label{eq:dufomap}
\end{equation}
where $\mathcal{R}(p_{t,i})$ is the DUFOMap~\cite{duberg2024dufomapefficientdynamicawareness} ray-casting projection operator. Dynamic points are extracted as $\mathcal{P}_t^d = \{p_{t,i} \mid \psi(p_{t,i}) = 1\}$. DBSCAN~\cite{ester1996density} is then applied to $\mathcal{P}_t^d$ to cluster dynamic objects.
Clusters are refined through multi-frame geometric consensus:
\begin{equation}
\hat{\mathcal{C}}_k^{(t)} = \mathcal{C}_k^{(t)} \cap \bigg(\bigcup_{\delta = -1, 1} \mathcal{C}_k^{(t + \delta)}\bigg)
\label{eq:temporal_refine}
\end{equation}
where \(\mathcal{C}_k^{(t)}\) refers to the cluster at time \(t\), and the temporal refinement combines the cluster information from neighboring frames \(t-1\), \(t\), and \(t+1\) to improve label consistency.
\subsubsection{Refined Instance Segmentation}
To enhance segmentation (Step 2 in \cref{alg:coarse_to_fine}), we use a voxel-based representation to capture local geometric features and apply a GRU-based refinement for feature enhancement. This approach improves clustering and segmentation by refining point cloud representations via voxel-wise feature aggregation and spatio-temporal refinement.

Given a point $p_{t,i} = (x_{t,i}, y_{t,i}, z_{t,i}) \in \mathbb{R}^3$ at frame $t$, its voxel coordinates are computed as:
\begin{equation}
v = \left( \left\lfloor \frac{x_{t,i}}{\Delta} \right\rfloor, \left\lfloor \frac{y_{t,i}}{\Delta} \right\rfloor, \left\lfloor \frac{z_{t,i}}{\Delta} \right\rfloor \right)
\label{eq:voxel}
\end{equation}
where $\Delta$ is the voxel size (default 0.3m). This ensures points in the same voxel share features while preserving spatial structure.

As detailed in Step 3 of \cref{alg:coarse_to_fine}, the GRU module refines features iteratively over $k$ steps:
\begin{equation}
\begin{aligned}
\mathbf{h}_{v,t}^{(k+1)} &= \text{GRU}_v \left( \mathbf{h}_{v,t}^{(k)}, \frac{1}{|v|} \sum_{i \in v} \mathbf{f}_{t,i}^{(k)} \right) \quad \forall v \in \mathcal{V}_t \\
\mathbf{f}_{t,i}^{(k+1)} &= \text{GRU}_p \left( \mathbf{f}_{t,i}^{(k)}, \mathbf{h}_{v,t}^{(k+1)} \right)
\end{aligned}
\label{eq:gru_update}
\end{equation}
where $\mathbf{h}_{v,t}^{(k)}$ is the hidden state of voxel $v$ at iteration $k$, and $\mathbf{f}_{t,i}^{(k)}$ is the refined feature of point $i$ after $k$ iterations.

\subsection{Mutually Promotion Loss Functions}
\label{subsec:loss}
While Chamfer loss in SeFlow \cite{zhang2024seflow} ensures point-wise registration, it has three key limitations in multi-task learning: (1) Insensitivity to instance-level motion coherence, (2) Lack of geometric constraints at segmentation boundaries, and (3) Over-smoothing objects that should remain distinct. To address these issues, we introduce a set of four loss functions that are both mutually reinforcing and constraining.

\begin{equation}
\mathcal{L}_{\text{total}} = \lambda_{\text{BF}}\mathcal{L}_{\text{BF}} + \lambda_{\text{Rigid}}\mathcal{L}_{\text{Rigid}} + \lambda_{\text{SMC}}\mathcal{L}_{\text{SMC}} + \lambda_{\text{DOM}}\mathcal{L}_{\text{DOM}}
\end{equation}
\subsubsection{Background-Foreground Segmentation Loss}
\label{sssec:focal_loss}

Given instance pseudo-labels \( \hat{l}_i \in \{0, 1, \ldots, C\} \), the binary mask is generated as \( y_i = H(\hat{l}_i) \), where \( H(x) \) is the Heaviside step function.
For the predicted logits \( \mathbf{m}_i \) with background channels \( \mathbf{m}_i^{(0)} \), the probabilities of background and foreground are computed as:
\begin{equation}
\begin{aligned}
\text{p}^{\text{bg}}_i &= \sigma(\mathbf{m}_i^{(0)}) \\
\text{p}^{\text{fg}}_i &= 1 - \text{p}^{\text{bg}}_i
\end{aligned}
\end{equation}
where \( \sigma(x) = \frac{1}{1 + e^{-x}} \) is the sigmoid function that maps logits to probabilities.
The balanced focal loss combines both foreground and background terms:
\begin{equation}
\mathcal{L}_{\text{BF}} = \frac{1}{N} \sum_{i=1}^N \left[ \beta \mathcal{F}(\text{p}^{\text{fg}}_i, y_i) + (1-\beta) \mathcal{F}(\text{p}^{\text{bg}}_i, 1 - y_i) \right],
\end{equation}
where the focal term is:
\begin{equation}
\mathcal{F}(\text{p}, y) = -\alpha y (1 - \text{p})^\gamma \log \text{p}
\end{equation}
and \( \alpha \), \( \gamma \), and \( \beta \) are hyperparameters that control the loss scale, focus on hard examples, and foreground-background weighting, respectively.


\subsubsection{Rigid Object Consistency Loss}
\label{sssec:rigid_loss}
\begin{figure}[ht]
  \centering
   \includegraphics[width=1.0\linewidth]{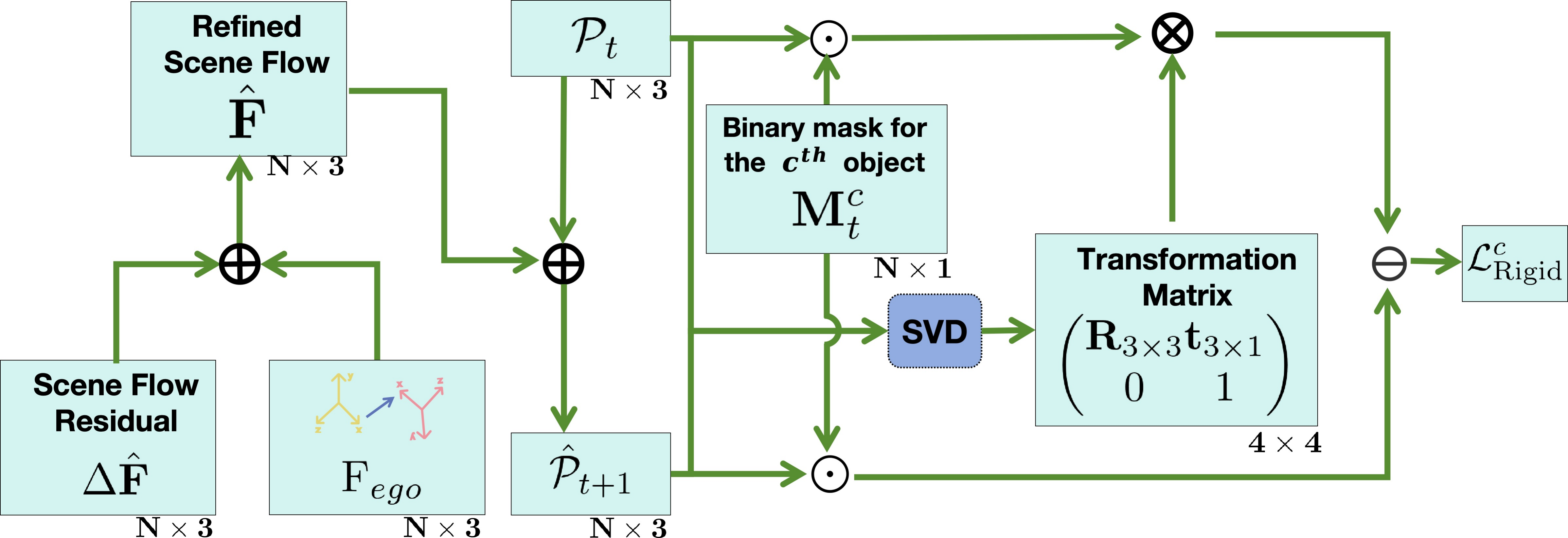}
   \caption{An illustration of $\mathcal{L}_{\text{Rigid}}$. The operator $\odot$, $\oplus$, and $\ominus$ denote element-wise multiplication, addition, and subtraction respectively, and $\otimes$ denotes matrix multiplication.}
   \label{fig:l_rigid}
\end{figure}
While SeFlow~\cite{zhang2024seflow} assumes uniform motion within clusters, this assumption breaks down for objects undergoing rotation or non-linear transformations. For example, vehicles turning on curved roads exhibit point motions in varying directions relative to the rotational center. As shown in \cref{fig:l_rigid}, we enforce consistent rigid body motion between consecutive frames.
\begin{equation}
\begin{aligned}
\mathcal{L}_{\text{Rigid}} = & \frac{1}{N*C} \sum_{c=1}^{C} \left\| 
    \left( \mathcal{P}_t \odot \mathbf{M}_t^{c} \right)\cdot\mathbf{R}^T \right. \\
    & \quad + 1_N \cdot \mathbf{t} - \left. \left( \hat{\mathcal{P}}_{t+1} \odot \mathbf{M}_{t}^{c} \right) \right\|_2,
\end{aligned}
\end{equation}
where
\( \mathbf{R} \) and \( \mathbf{t} \) are the optimal rotation matrix and translation vector, obtained through Singular Value Decomposition (SVD)\cite{abdi2007singular} based method \cite{kabsch1976solution,gojcic2020learning} during rigid body alignment, \( \mathbf{M}_t^{c} \) is the binary mask  for the object \( c \).


\subsubsection{Spatial Mask Consistency Loss}
\label{sssec:smc_loss}
To enforce similarity between each point's mask and those of its neighbors, the loss is defined as:

\begin{equation}
\begin{aligned}
\mathcal{L}_{\text{SMC}} = \frac{1}{N} \sum_{i=1}^N \Big( 
    & w_{\text{KNN}} \sum_{j \in \mathcal{N}_k(i)} \| \textbf{m}_i - \textbf{m}_j \|_2^2 \\
    & + w_{\text{BallQ}} \sum_{j \in \mathcal{B}_r(i)} \| \textbf{m}_i - \textbf{m}_j \|_2^2 
\Big),
\end{aligned}
\end{equation}
two neighborhood definitions, K-Nearest Neighbors (KNN) $\mathcal{N}_k(i)$ and Ball Query $\mathcal{B}_r(i)$, are employed, with weights $w_{\text{KNN}}$ and $w_{\text{BallQ}}$ balancing their contributions.


\subsubsection{Dynamic Object Mask Loss
}
\label{sssec:dom_loss}
To address the over-smoothing caused by the $\mathcal{L}_{\text{SMC}}$, this loss penalizes excessive similarity between the logits of spatially distant points:
\begin{equation}
\mathcal{L}_{\text{DOM}} = 
\begin{cases} 
\frac{1}{|\mathcal{P}|} \sum_{(i, j) \in \mathcal{P}} \frac{\mathbf{m}_i \cdot \mathbf{m}_j}{\|\mathbf{m}_i\|_2 \|\mathbf{m}_j\|_2}, & \text{if } |\mathcal{P}| > 0, \\
\mathcal{L}_{\text{min}}, & \text{otherwise},
\end{cases}
\end{equation}
where \(\mathcal{P} = \{(i, j) \mid \|p_{t,i} - p_{t,j}\|_2 > \delta\}\) represents pairs of foreground points whose Euclidean distance exceeds a threshold \(\delta\). If no such pairs exist (\(|\mathcal{P}| = 0\)), the loss defaults to a predefined minimum value \(\mathcal{L}_{\text{min}}=1.0\). 

\begin{algorithm}[t]
\caption{Joint Self-Supervised Learning for Scene Flow Estimation and Segmentation}
\label{alg:pipeline}
\KwIn{$\mathcal{P}_t, \mathcal{P}_{t+1} \in \mathbb{R}^{N \times 3}$}
\KwOut{$\mathcal{L}_{total}$: total loss for backward}

\textbf{Dynamic Clustering:} Generate $\mathcal{C}_k^{(t)}$ via \cref{eq:temporal_refine} \;
$\mathcal{P}_t^s = \{ p_{t,i} \mid p_{t,i} \notin \bigcup\limits_{k} \mathcal{C}_k^{(t)} \}$\;
$\mathbf{F}_\text{ego} \gets \text{ICP}(\mathcal{P}_t^s, \mathcal{P}_{t+1}^s)$ \cite{lin2024icp} \;
\For{\text{epoch} = 1 \textbf{to} $E$}{
    $\hat{l} \gets \{\}$\;
    $\Delta\hat{\mathbf{F}} \gets \{\}$\;
    \For{\text{each } $p_{t,i} \in \mathcal{P}_t$}{
    $\mathbf{h}_{v,t}^{(K)}, \mathbf{f}_{t,i}^{(K)} \gets \text{obtained via \cref{alg:coarse_to_fine}}$ \;
    
    $\mathbf{f}_{t,i}^{\text{shared}} \gets \text{MLP}([\mathbf{f}_{t,i}^{(0)}; \mathbf{f}_{t,i}^{(K)}])$ \;
    $\Delta\hat{\mathbf{F}} _{t,i}\gets w_{\text{flow}} \mathbf{f}_{t,i}^{\text{shared}}$ \;
    $\mathbf{m}_{i}\gets \text{Softmax}(w_{\text{seg}} \mathbf{f}_{t,i}^{\text{shared}})$\;
    $\hat{l}_i \gets \arg\max(\mathbf{m}_i)$\;
    $\hat{l} \gets \hat{l} \cup \{\hat{l}_i\}$\;
    $\Delta\hat{\mathbf{F}} \gets \Delta\hat{\mathbf{F}} _{t} \cup \{\Delta\hat{\mathbf{F}} _{t,i}\}$
    }

    $\mathbf{M}_t \gets [\mathbf{m}_1, \dots, \mathbf{m}_N]^\top$ \;
    $\hat{\mathbf{F}} \gets\mathbf{F}_\text{ego} + \Delta \hat{\mathbf{F}}$ \;
    
     $\mathcal{L}_{\text{BF}} \gets\text{using $\hat{l}$ and $\mathbf{M}_t$ via \cref{sssec:focal_loss}}$\;

     $\mathcal{L}_{\text{Rigid}} \gets\text{using $\mathcal{P}_t$, $\mathbf{M}_t$,$\hat{\mathbf{F}}$ via \cref{sssec:rigid_loss}}$\;
    
     $\mathcal{L}_{\text{SMC}} \gets\text{using $\mathcal{P}_t$ and $\mathbf{M}_t$ via \cref{sssec:smc_loss}}$\;

     $\mathcal{L}_{\text{DOM}} \gets\text{using $\mathcal{P}_t$ and $\mathbf{M}_t$ via \cref{sssec:dom_loss}}$
}
\end{algorithm}
\subsection{Self-Supervised Learning Process}
\label{subsec:architecture}

The self-supervised learning process, including feature refinement and optimization, is outlined in \cref{alg:pipeline}. Clustering identifies static and dynamic regions, enabling ego-motion estimation via ICP \cite{lin2024icp}. Each point undergoes voxel-based feature refinement through a recurrent update scheme, with refined features aggregated for scene flow residual and instance segmentation predictions. The instance mask and scene flow are guided by self-supervised losses enforcing motion consistency, segmentation accuracy, and spatial coherence.
The relationships between loss functions are crucial:
(1) \( \mathcal{L}_{\text{Rigid}} \) and \( \mathcal{L}_{\text{BF}} \) constrain each other: \( \mathcal{L}_{\text{Rigid}} \) minimizes scene flow, which may lead to treating all points as a single motion, while \( \mathcal{L}_{\text{BF}} \) enforces background-foreground segmentation, ensuring dynamic object separation.
(2) \( \mathcal{L}_{\text{SMC}} \) and \( \mathcal{L}_{\text{DOM}} \) also impose mutual constraints: \( \mathcal{L}_{\text{SMC}} \) discourages excessive segmentation, preserving real-world object boundaries, while \( \mathcal{L}_{\text{DOM}} \) prevents foreground merging due to overly strict segmentation, ensuring spatial coherence.
The training framework optimizes segmentation for improved scene flow estimation, with each loss function guiding the others.

\section{Experiments}
\begin{table*}[!t]\scriptsize
  \centering
  \begin{threeparttable}[c]
    \renewcommand\arraystretch{1.5}
    \setlength{\tabcolsep}{10pt}
    \caption{Scene Flow Estimation Accuracy on Waymo validation set and Argoverse 2 test set (measured by EPE$\downarrow$)}
    \label{tab:Scene_Flow_Prediction_Accuracy}
    \begin{tabular}{c|C{0.9cm}|C{0.9cm}|cccc|cccc}
    \Xhline{2pt}
    \multirow{2}{*}{Method} &\multirow{2}{*}{Supervised} & \multirow{2}{*}{Multi-task} & \multicolumn{4}{c|}{$Waymo$ (validation set)} & \multicolumn{4}{c}{$Argoverse2$ (test set)} \\
    \cline{4-11}
    & & & 3-way & BS & FS & FD & 3-way & BS & FS & FD \\
    \cline{1-11}
    FastFlow3D\cite{jund2021scalablesceneflowpoint} & Yes & No & 0.0784 & 0.0152 & 0.0246 & 0.1954 & 0.0620 & 0.0049 & 0.0245 & 0.1564 \\
    DeFlow\cite{zhang2024deflow} & Yes & No & 0.0446 & 0.0098 & 0.0259 & 0.0980 & 0.0343 & 0.0046 & 0.0251 & 0.0732 \\
    \cdashline{1-11}[2pt/2pt]
    FastNSF\cite{li2023fast} & No & No & 0.1579 & 0.0403 & 0.0146 & 0.3012 & 0.1118 & 0.0907 & 0.0814 & 0.1634 \\
    ZeroFlow\cite{vedder2023zeroflow} & No & No & 0.0921 & 0.0241 & 0.0153 & 0.2162 & 0.0494 & 0.0131 & 0.0174 & \textbf{0.1177} \\
    ICPFlow\cite{lin2024icp} & No & No & 0.1176 & 0.0830 & 0.0865 & 0.1834 & 0.0650 & 0.0250 & 0.0332 & 0.1369 \\
    SeFlow\cite{zhang2024seflow} & No & No & 0.0598 & 0.0106 & 0.0181 & 0.1506 & 0.0486 & 0.0060 & 0.0184 & 0.1214 \\
    \cdashline{1-11}[2pt/2pt]
    SemanticFlow (Ours) & No & \textbf{Yes} & \textbf{0.0545} & \textbf{0.0080} & \textbf{0.0122} & \textbf{0.1433} & \textbf{0.0469} & \textbf{0.0040} & \textbf{0.0141} & 0.1226 \\
    \bottomrule[2pt]
    \end{tabular}
  \end{threeparttable}
\end{table*}
\begin{table*}[!t]\scriptsize
  \centering
  \begin{threeparttable}[c]
    \renewcommand\arraystretch{1.5}
    \setlength{\tabcolsep}{10pt}
    \caption{Segmentation Accuracy on Argoverse 2 validation set}
    \label{tab:Segmentation_Accuracy}
    \begin{tabular}{c|C{0.9cm}|C{0.9cm}|ccccccc}
    \Xhline{2pt}
    \multirow{2}{*}{Method} &\multirow{2}{*}{Supervised}& \multirow{2}{*}{Multi-task}  & \multicolumn{7}{c}{$Argoverse2$ (validation set)} \\
    \cline{4-10}
    & & & AP$\uparrow$ & PQ$\uparrow$ & F1$\uparrow$ & Pre$\uparrow$ & Rec$\uparrow$ & mIoU$\uparrow$ & RI$\uparrow$\\
    \cline{1-10}
    $\text{OGC}_{sup}$\cite{NEURIPS2022_c6e38569} & Yes & No & 61.3 & 53.4 & 63.8 & 60.0 & 64.2 & 65.5 & 88.0 \\
    $\text{SemanticFlow}_{sup}$(Ours) & Yes & No & 63.5 & 57.8 & 65.9 & 61.4 & 65.3 & 69.8 & 94.3  \\
    \cdashline{1-10}[2pt/2pt]
    WardLinkage\cite{ward1963hierarchical} & No & No & 25.7 & 12.5 & 25.2 & 10.1 & 57.7 & 58.3 & 41.7 \\
    DBSCAN\cite{ester1996density}
  & No & No & 11.7 & 23.9 & 15.3 & 21.6 & 49.4 & 43.1 & 57.2 \\
    \cdashline{1-10}[2pt/2pt]
    OGC\cite{NEURIPS2022_c6e38569} & No & No & 51.4 & 48.0 & 51.2 & 44.3 & 54.6 & 60.4 & 81.4\\
    SemanticFlow (Ours) & No & \textbf{Yes} & \textbf{57.5} & \textbf{50.8} & \textbf{52.2} & \textbf{54.3} & \textbf{64.6} & \textbf{64.0} & \textbf{91.4}\\
    \bottomrule[2pt]
    \end{tabular}
  \end{threeparttable}
\end{table*}
\begin{figure*}[!t]
    \centering
    \begin{tabular}{cccc}
        \begin{subfigure}[b]{0.22\textwidth}
            \includegraphics[width=\textwidth]{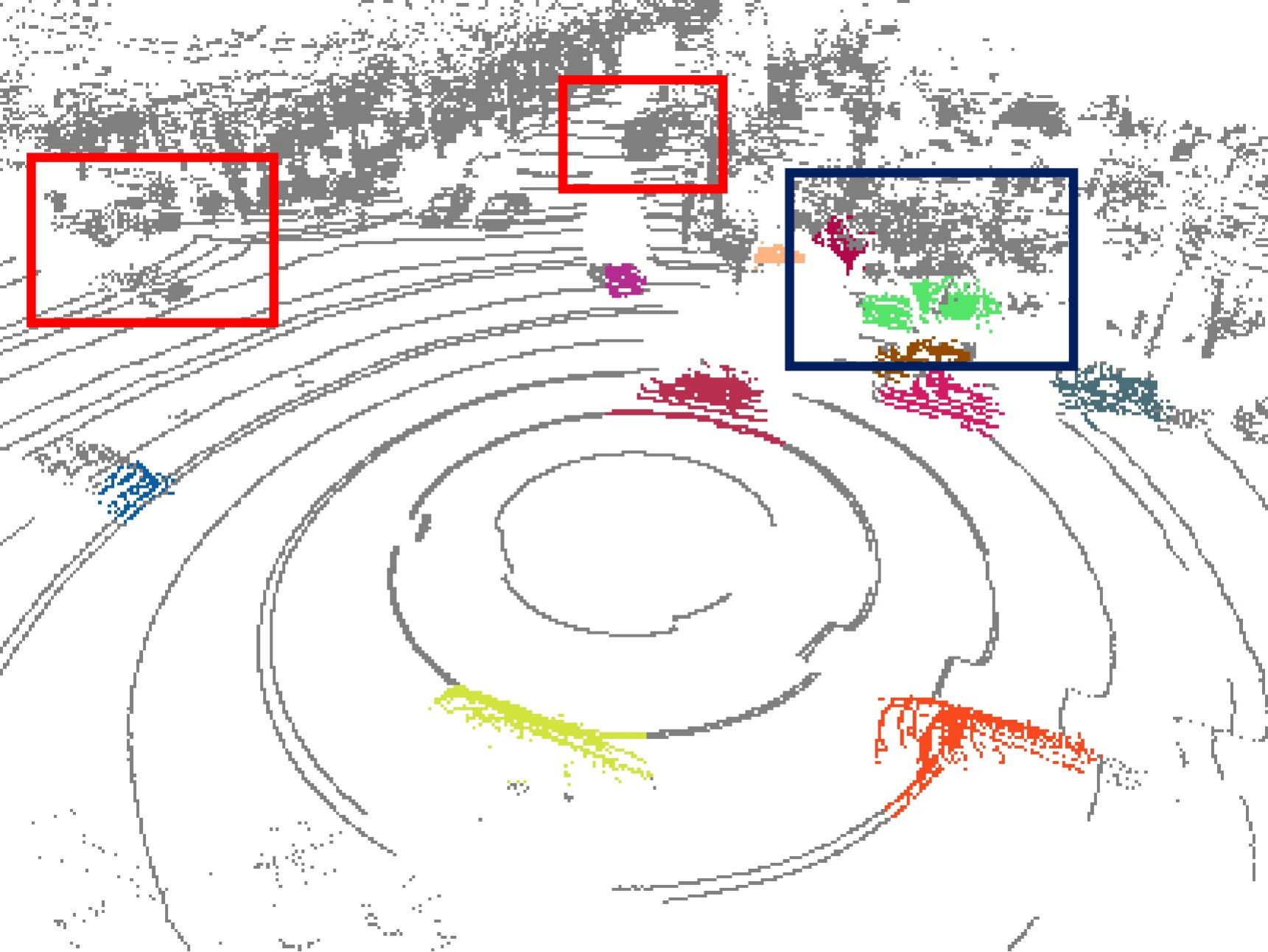}
            \caption{\textit{WardLinkage}}
        \end{subfigure} &
        \begin{subfigure}[b]{0.22\textwidth}
            \includegraphics[width=\textwidth]{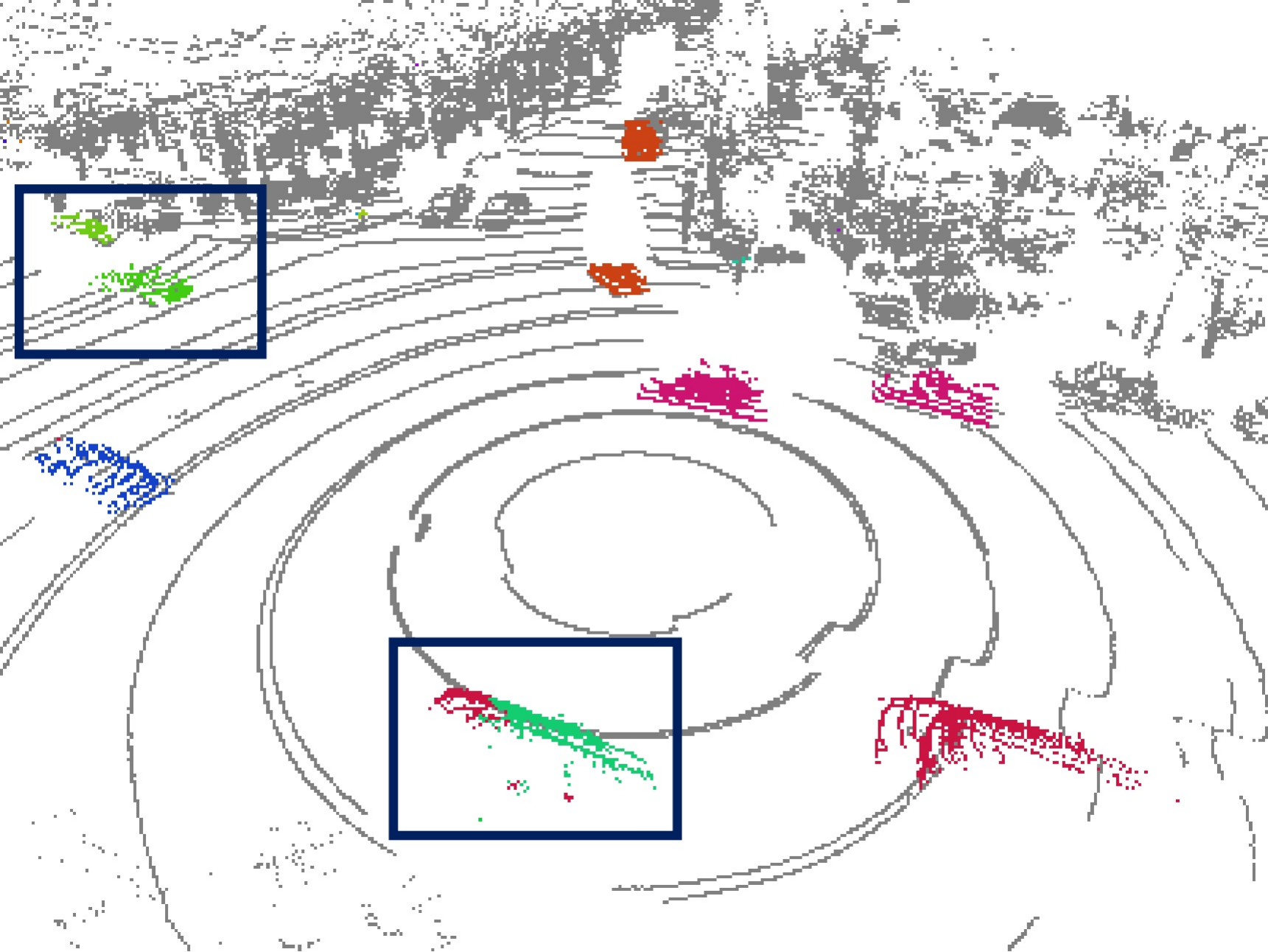}
            \caption{\textit{OGC$_{sup}$}}
        \end{subfigure} &
        \begin{subfigure}[b]{0.22\textwidth}
            \includegraphics[width=\textwidth]{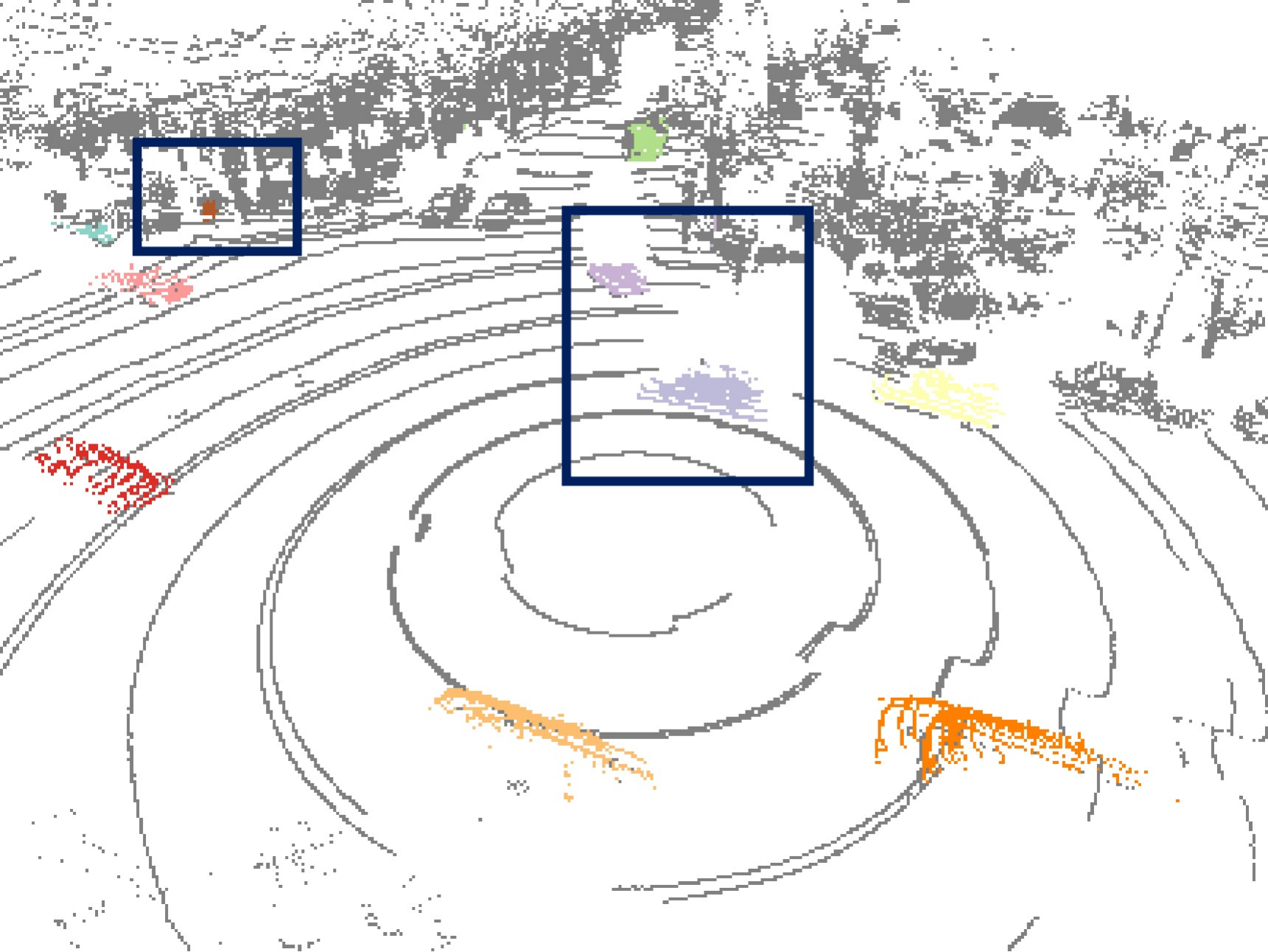}
            \caption{\textit{OGC}}
        \end{subfigure} &

        \multirow{2}{*}[25mm]{\begin{adjustbox}{valign=t}
            \begin{subfigure}[b]{0.22\textwidth}
                \includegraphics[width=\textwidth]{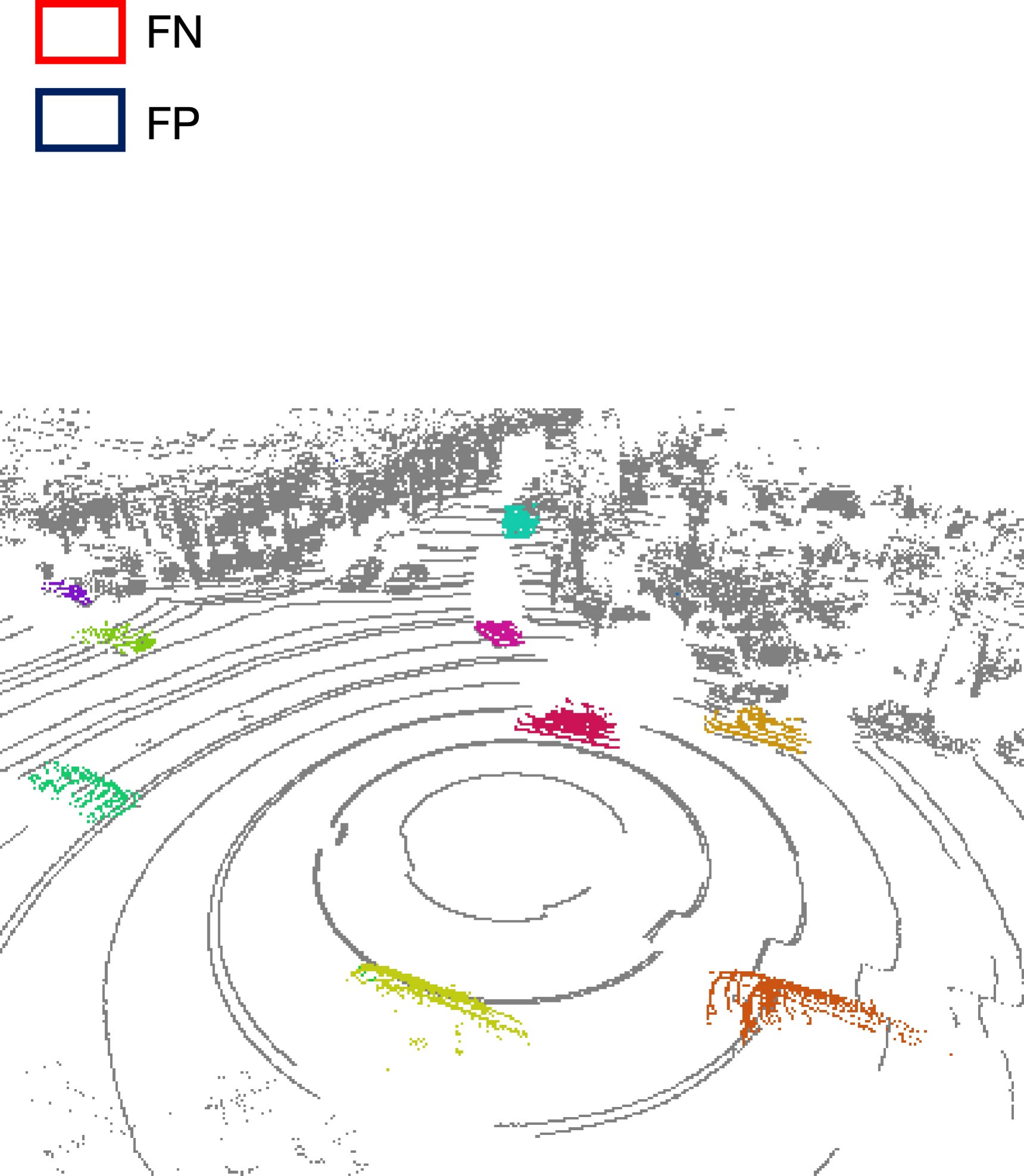}
                \caption{\textit{Ground Truth}}
            \end{subfigure}
        \end{adjustbox}} \\ 
        
        \begin{subfigure}[b]{0.22\textwidth}
            \includegraphics[width=\textwidth]{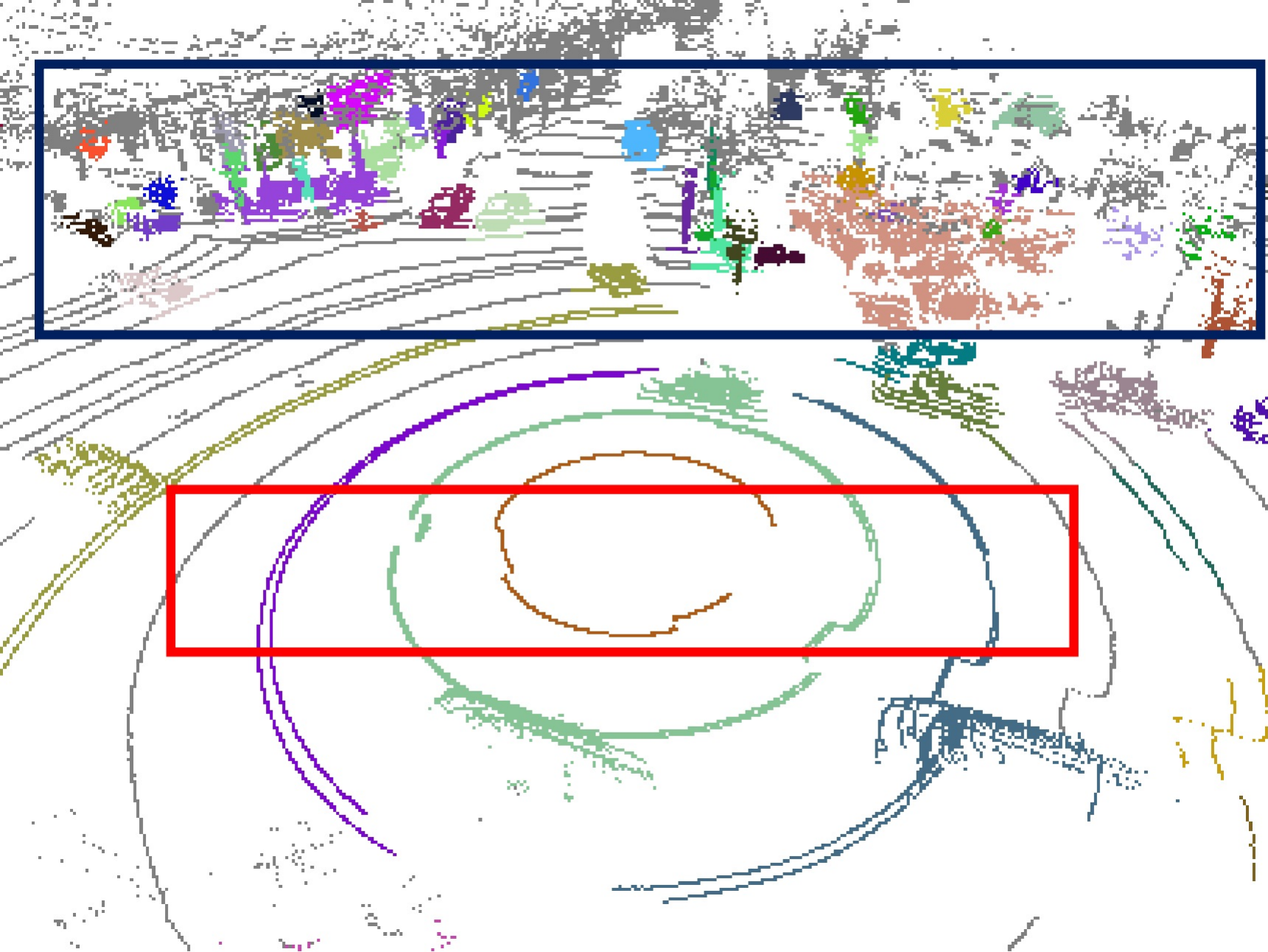}
            \caption{\textit{DBSCAN}}
        \end{subfigure} &
        \begin{subfigure}[b]{0.22\textwidth}
            \includegraphics[width=\textwidth]{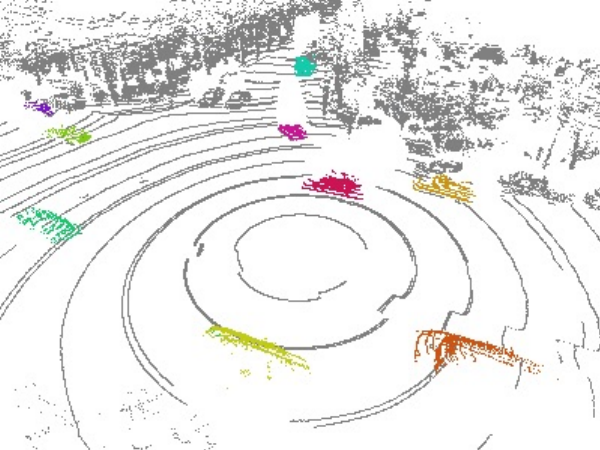}
            \caption{\textit{SemanticFlow$_{sup}$(Ours)}}
        \end{subfigure} &
        \begin{subfigure}[b]{0.22\textwidth}
            \includegraphics[width=\textwidth]{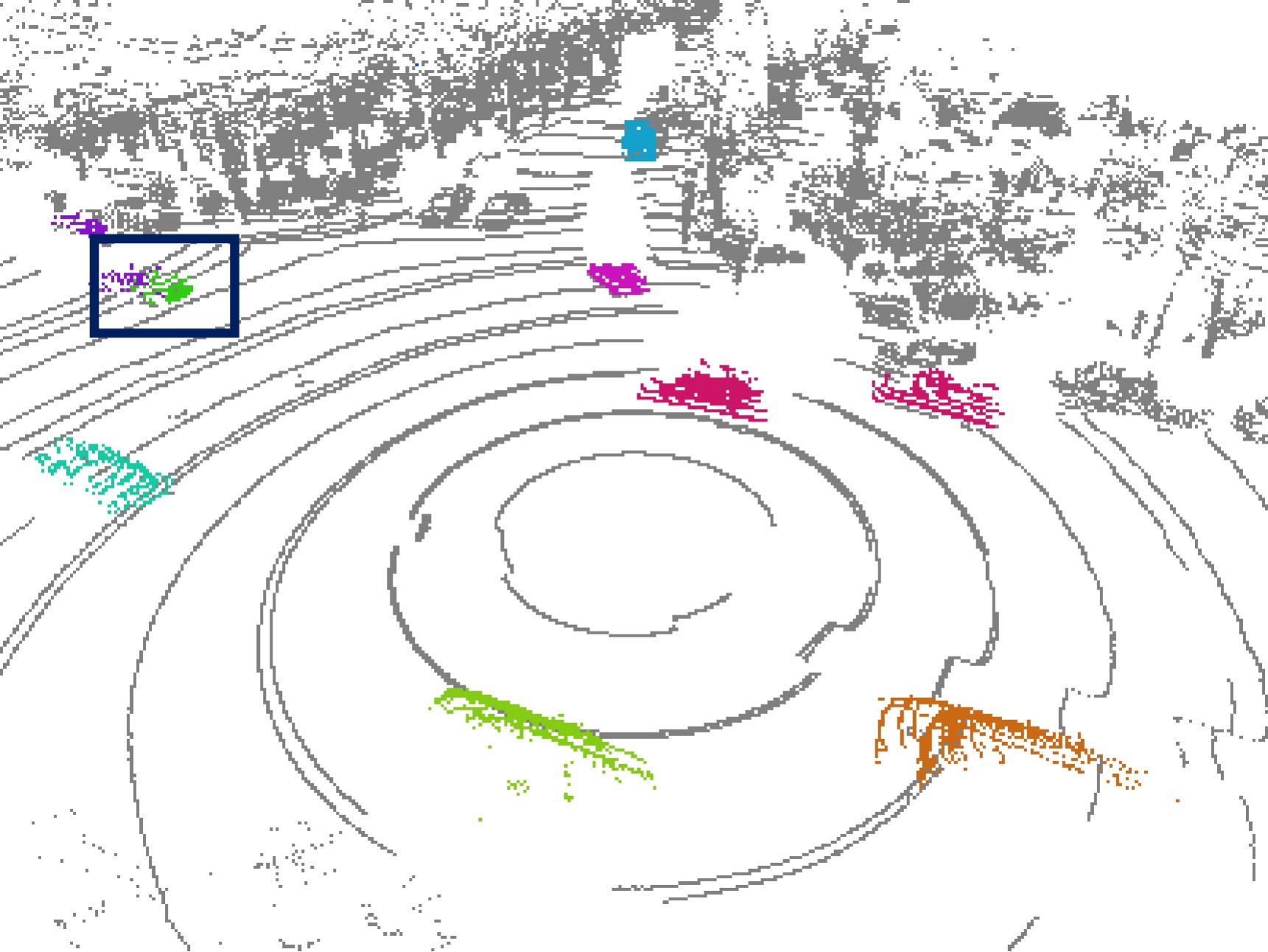}
            \caption{\textit{SemanticFlow (Ours)}}
        \end{subfigure} & \\ 
    \end{tabular}
    
    \caption{Comparison of different clustering methods for scene flow segmentation. False negatives (FN) are marked in \textcolor{red}{red}, and false positives (FP) are marked in \textcolor{blue}{blue}.}
    \label{fig:compare}
\end{figure*}

\subsection{Validation of SemanticFlow on Public Datasets}

\subsubsection{Datasets}
We validate the performance of SemanticFlow on two widely-used large-scale datasets: Waymo\cite{sun2020scalability} and Argoverse 2 \cite{wilson2023argoverse}. The Waymo dataset comprises 798 training scenes and 202 validation scenes, capturing diverse driving scenarios. Argoverse 2 provides 700 training scenes, 150 validation scenes, and 150 test scenes, with test results evaluated via a public leaderboard. Additionally, Argoverse 2 includes a LiDAR dataset with 20,000 unlabeled scenes, facilitating unsupervised learning and testing the adaptability of self-supervised methods in realistic settings.
\subsubsection{Metrics}
The evaluation metric for scene flow is the three-way End Point Error (3-way EPE), calculated as the L2 norm of the difference between predicted and ground-truth flow vectors (in meters). The 3-way EPE is divided into three subsets: Background Static (BS), Foreground Static (FS), and Foreground Dynamic (FD). A point is classified as dynamic if its flow magnitude exceeds 0.05m (0.5 m/s). Evaluations are performed within a 100m × 100m region centered on the ego vehicle for consistency.

Segmentation performance is evaluated using metrics such as Average Precision (AP), Panoptic Quality (PQ), F1 score, Precision (Pre), Recall (Rec), mean Intersection over Union (mIoU), and Rand Index (RI), with higher values indicating better performance.


\subsubsection{Comparative Results}
\cref{tab:Scene_Flow_Prediction_Accuracy} compares SemanticFlow with other methods on the Waymo validation and Argoverse 2 test sets. Supervised approaches like FastFlow3D~\cite{jund2021scalablesceneflowpoint} and DeFlow\cite{zhang2024deflow} benefit from labeled data for strong performance, while our self-supervised SemanticFlow achieves comparable results with the added advantage of instance segmentation.  
SemanticFlow stands out by maintaining competitive 3-way EPE values while outperforming in FS and BS metrics, demonstrating its superior ability to distinguish dynamic and static objects, which is crucial for downstream perception tasks. On the Waymo dataset, it improves 3-way EPE by $9\%$ over SeFlow (0.0545 vs. 0.0598), and on the Argoverse 2 test set, it achieves slightly better accuracy (0.0469 vs. 0.0486). Despite incorporating instance segmentation in a multi-task framework, SemanticFlow maintains high scene flow estimation accuracy, offering richer, more actionable insights for real-world applications.  

\cref{tab:Segmentation_Accuracy} presents the quantitative results of our method, SemanticFlow, alongside several baseline methods on the Argoverse 2 validation set. While supervised methods, such as \(\text{OGC}_{sup}\) and \(\text{SemanticFlow}_{sup}\), achieve strong results, our self-supervised SemanticFlow, which incorporates multi-task learning, outperforms all other methods in most metrics. Specifically, SemanticFlow achieves significant improvements in \textbf{AP} (57.5 vs. 51.4) and \textbf{PQ} (50.8 vs. 48.0) compared to the baseline method OGC, as well as higher \textbf{mIoU} (64.0 vs. 60.4) and \textbf{RI} (91.4 vs. 81.4) compared to other unsupervised methods.

\cref{fig:compare} illustrates that the multi-task framework of SemanticFlow enhances both segmentation accuracy and consistency, surpassing methods that do not utilize multi-task learning, such as \(\text{OGC}\), as well as those based on unsupervised clustering techniques like DBSCAN and WardLinkage. The integration of instance segmentation with scene flow estimation further contributes to the superior performance of SemanticFlow, making it a promising solution for real-world dynamic scene understanding.

\begin{table*}[!t]\scriptsize
  \centering
  \begin{threeparttable}[c]
    \renewcommand\arraystretch{1.5}
    \setlength{\tabcolsep}{10pt}
    \caption{Ablation Study on Loss Components for Scene Flow Estimation}
    \label{tab:Ablation_Study}
    \begin{tabular}{C{0.2cm}|C{0.4cm}C{0.4cm}C{0.4cm}C{0.4cm}C{0.4cm}|cccc|cccc}
    \Xhline{2pt}
    \multirow{2}{*}{ID} & \multicolumn{5}{c|}{$Losses$} & \multicolumn{4}{c|}{$Waymo$ (validation set)} & \multicolumn{4}{c}{$Argoverse2$ (test set)} \\
    \cline{2-14} &
    $\mathcal{L}_{\text{Seflow}}$ & $\mathcal{L}_{\text{BF}}$ & $\mathcal{L}_{\text{Rigid}}$ & $\mathcal{L}_{\text{SMC}}$ & $\mathcal{L}_{\text{DOM}}$ & BS & FS & FD & 3-way & BS & FS & FD & 3-way \\
    \cline{1-14}
    1 & \CheckmarkBold &  &  & & & 0.0106 & 0.0181 & 0.1506 & 0.0598 & 0.0060 & 0.0184 & 0.1214 & 0.0486 \\
    2 & \CheckmarkBold & \CheckmarkBold &  & & & 0.0104 & 0.0166 & 0.1498 & 0.0589 & 0.0056 & 0.0180 & 0.1432 & 0.0556 \\
    3 & \CheckmarkBold & \CheckmarkBold & \CheckmarkBold &  & & 0.0093 & 0.0154 & 0.1467 & 0.0571 & 0.0053 & 0.0163 & 0.1296 & 0.0504 \\
    4 & \CheckmarkBold & \CheckmarkBold & \CheckmarkBold & \CheckmarkBold & & 0.0082 & 0.0130 & 0.1450 & 0.0554 & 0.0044 & 0.0143 & 0.1278 & 0.0488 \\
    5 & \CheckmarkBold & \CheckmarkBold & \CheckmarkBold & \CheckmarkBold & \CheckmarkBold &  \textbf{0.0080} & \textbf{0.0122} & \textbf{0.1433} & \textbf{0.0545} & \textbf{0.0040} & \textbf{0.0141} & 0.1226 & \textbf{0.0469} \\
    \bottomrule[2pt]
    \end{tabular}
  \end{threeparttable}
\end{table*}

\subsection{Ablation Study: Analysis of Loss Components}

To quantify the contributions of individual loss components, we conducted ablation studies on the Waymo validation set and the Argoverse2 test set. All experiments were trained for 20 epochs under a consistent setup.

The baseline model, which relies solely on the scene flow loss $\mathcal{L}_{\text{Seflow}}$ from Seflow\cite{zhang2024deflow}, struggles to capture the distinct motion characteristics of static and dynamic regions, leading to suboptimal performance. For instance, in the Waymo dataset, the EPE of BS is 0.0106, FS is 0.0181, and FD is 0.1506, highlighting the model’s difficulty in distinguishing background from foreground and in handling dynamic object motion.

Introducing the background-foreground separation loss $\mathcal{L}_{\text{BF}}$ (Experiment 2) enhances structural differentiation between static and dynamic regions. While improvements are observed in BS and FS across both datasets, FD shows only marginal gains, suggesting that $\mathcal{L}_{\text{BF}}$ mainly benefits static region segmentation but is insufficient for refining dynamic object motion.

The rigid consistency loss $\mathcal{L}_{\text{Rigid}}$ (Experiment 3) significantly improves dynamic foreground estimation by enforcing motion coherence within rigid bodies. EPE of FD decreases notably (e.g., 9.5\% reduction in Argoverse2), demonstrating that explicitly modeling rigid motion constraints strengthens flow consistency for dynamic objects.

The spatial mask consistency loss $\mathcal{L}_{\text{SMC}}$ (Experiment 4) further enhances spatial coherence in both segmentation and flow prediction. Notably, FS improves substantially (e.g., 12.3\% in Argoverse2), underscoring its role in refining the segmentation of static foreground objects and mitigating spatial inconsistencies in learned representations.

Integrating the dynamic object mask $\mathcal{L}_{\text{DOM}}$ (Experiment 5) addresses over-smoothing in self-supervised foreground segmentation, ensuring sharper boundaries and improved motion estimation. As a result, both segmentation and flow estimation reach optimal accuracy, with EPE of FD  decreasing by 4.1\% in Argoverse 2. This highlights $\mathcal{L}_{\text{DOM}}$ as a key component for refining motion segmentation and improving dynamic objects' scene flow estimation.

In summary, the full model incorporating all loss components achieves the most comprehensive improvement. Compared to the baseline, EPE of BS in Waymo reduces by 24.5\%, and FD decreases by 5.5\%. In Argoverse 2, BS improves by 33.3\%, and FD by 14.4\%. These results confirm that each loss function plays a distinct yet complementary role: $\mathcal{L}_{\text{BF}}$ enhances foreground-background separation, $\mathcal{L}_{\text{Rigid}}$ strengthens rigid motion consistency, $\mathcal{L}_{\text{SMC}}$ ensures spatial coherence, and $\mathcal{L}_{\text{DOM}}$ optimizes dynamic object segmentation. Together, these components enable a robust and accurate self-supervised framework for joint segmentation and scene flow estimation.
\section{CONCLUSION}

In this paper, we introduced SemanticFlow, a self-supervised multi-task learning framework 
for joint scene flow estimation and instance segmentation. By leveraging innovative loss functions, SemanticFlow effectively addresses the challenges of dynamic scene understanding, enabling accurate scene flow predictions and robust segmentation of both dynamic and static objects. Extensive experiments on Waymo and Argoverse 2 datasets demonstrate that SemanticFlow outperforms existing methods
, achieving a 8.7 $\sim$ 65.5\% improvement** in scene flow estimation and a 3.6 $\sim$ 11.9\% increase in segmentation accuracy** across key metrics, including 3-way EPE, foreground segmentation (FS), and background segmentation (BS). Our results highlight the competitive performance of SemanticFlow against supervised methods, reinforcing its potential for real-world applications. The integration of multi-task learning enhances its practical utility for downstream tasks such as SLAM, obstacle detection, decision planning, and tracking. 
This work showcases the effectiveness of self-supervised learning in complex dynamic scenes, with the potential for further improvements by exploring additional self-supervised techniques. 
Future efforts will focus on scaling the approach and evaluating its performance in more challenging real-world environments.


{
    \small
    \bibliographystyle{ieeenat_fullname}
    \bibliography{main}

\begin{thebibliography}{40}
\providecommand{\natexlab}[1]{#1}
\providecommand{\url}[1]{\texttt{#1}}
\expandafter\ifx\csname urlstyle\endcsname\relax
  \providecommand{\doi}[1]{doi: #1}\else
  \providecommand{\doi}{doi: \begingroup \urlstyle{rm}\Url}\fi

\bibitem[Abdi(2007)]{abdi2007singular}
Herv{\'e} Abdi.
\newblock Singular value decomposition (svd) and generalized singular value decomposition.
\newblock \emph{Encyclopedia of measurement and statistics}, 907\penalty0 (912):\penalty0 44, 2007.

\bibitem[Bahraini et~al.(2018)Bahraini, Bozorg, and Rad]{bahraini2018slam}
Masoud~S Bahraini, Mohammad Bozorg, and Ahmad~B Rad.
\newblock Slam in dynamic environments via ml-ransac.
\newblock \emph{Mechatronics}, 49:\penalty0 105--118, 2018.

\bibitem[Baur et~al.(2021)Baur, Emmerichs, Moosmann, Pinggera, Ommer, and Geiger]{Baur2021ICCV}
Stefan Baur, David Emmerichs, Frank Moosmann, Peter Pinggera, Bjorn Ommer, and Andreas Geiger.
\newblock Slim: Self-supervised lidar scene flow and motion segmentation.
\newblock In \emph{International Conference on Computer Vision (ICCV)}, 2021.

\bibitem[Chen et~al.(2024)Chen, Cui, Liu, Zhang, Sun, Ai, Gu, Xu, and Lu]{chen2024joint}
Xieyuanli Chen, Jiafeng Cui, Yufei Liu, Xianjing Zhang, Jiadai Sun, Rui Ai, Weihao Gu, Jintao Xu, and Huimin Lu.
\newblock Joint scene flow estimation and moving object segmentation on rotational lidar data.
\newblock \emph{IEEE Transactions on Intelligent Transportation Systems}, 2024.

\bibitem[Chen et~al.(2025)Chen, Zhang, Hao, and Zhou]{chen2025ssfpansemanticsceneflowbased}
Yinqi Chen, Meiying Zhang, Qi Hao, and Guang Zhou.
\newblock Ssf-pan: Semantic scene flow-based perception for autonomous navigation in traffic scenarios, 2025.

\bibitem[Crawshaw(2020)]{crawshaw2020multitasklearningdeepneural}
Michael Crawshaw.
\newblock Multi-task learning with deep neural networks: A survey, 2020.

\bibitem[Dey and Salem(2017)]{dey2017gate}
Rahul Dey and Fathi~M Salem.
\newblock Gate-variants of gated recurrent unit (gru) neural networks.
\newblock In \emph{2017 IEEE 60th international midwest symposium on circuits and systems (MWSCAS)}, pages 1597--1600. IEEE, 2017.

\bibitem[Duberg et~al.(2024)Duberg, Zhang, Jia, and Jensfelt]{duberg2024dufomapefficientdynamicawareness}
Daniel Duberg, Qingwen Zhang, MingKai Jia, and Patric Jensfelt.
\newblock Dufomap: Efficient dynamic awareness mapping, 2024.

\bibitem[Ester et~al.(1996)Ester, Kriegel, Sander, Xu, et~al.]{ester1996density}
Martin Ester, Hans-Peter Kriegel, J{\"o}rg Sander, Xiaowei Xu, et~al.
\newblock A density-based algorithm for discovering clusters in large spatial databases with noise.
\newblock In \emph{kdd}, pages 226--231, 1996.

\bibitem[Gojcic et~al.(2020)Gojcic, Zhou, Wegner, Guibas, and Birdal]{gojcic2020learning}
Zan Gojcic, Caifa Zhou, Jan~D Wegner, Leonidas~J Guibas, and Tolga Birdal.
\newblock Learning multiview 3d point cloud registration.
\newblock In \emph{Proceedings of the IEEE/CVF conference on computer vision and pattern recognition}, pages 1759--1769, 2020.

\bibitem[Gojcic et~al.(2021)Gojcic, Litany, Wieser, Guibas, and Birdal]{gojcic2021weakly}
Zan Gojcic, Or Litany, Andreas Wieser, Leonidas~J Guibas, and Tolga Birdal.
\newblock Weakly supervised learning of rigid 3d scene flow.
\newblock In \emph{Proceedings of the IEEE/CVF conference on computer vision and pattern recognition}, pages 5692--5703, 2021.

\bibitem[Huang et~al.(2021)Huang, Wang, Birdal, Sung, Arrigoni, Hu, and Guibas]{huang2021multibodysync}
Jiahui Huang, He Wang, Tolga Birdal, Minhyuk Sung, Federica Arrigoni, Shi-Min Hu, and Leonidas~J Guibas.
\newblock Multibodysync: Multi-body segmentation and motion estimation via 3d scan synchronization.
\newblock In \emph{Proceedings of the IEEE/CVF Conference on Computer Vision and Pattern Recognition}, pages 7108--7118, 2021.

\bibitem[Jund et~al.(2021)Jund, Sweeney, Abdo, Chen, and Shlens]{jund2021scalablesceneflowpoint}
Philipp Jund, Chris Sweeney, Nichola Abdo, Zhifeng Chen, and Jonathon Shlens.
\newblock Scalable scene flow from point clouds in the real world, 2021.

\bibitem[Kabir et~al.(2025)Kabir, Jim, and Istenes]{kabir2025terrain}
Md~Mohsin Kabir, Jamin~Rahman Jim, and Zolt{\'a}n Istenes.
\newblock Terrain detection and segmentation for autonomous vehicle navigation: A state-of-the-art systematic review.
\newblock \emph{Information Fusion}, 113:\penalty0 102644, 2025.

\bibitem[Kabsch(1976)]{kabsch1976solution}
Wolfgang Kabsch.
\newblock A solution for the best rotation to relate two sets of vectors.
\newblock \emph{Acta Crystallographica Section A: Crystal Physics, Diffraction, Theoretical and General Crystallography}, 32\penalty0 (5):\penalty0 922--923, 1976.

\bibitem[Kittenplon et~al.(2021)Kittenplon, Eldar, and Raviv]{kittenplon2021flowstep3d}
Yair Kittenplon, Yonina~C Eldar, and Dan Raviv.
\newblock Flowstep3d: Model unrolling for self-supervised scene flow estimation.
\newblock In \emph{Proceedings of the IEEE/CVF Conference on Computer Vision and Pattern Recognition}, pages 4114--4123, 2021.

\bibitem[Lentsch et~al.(2024)Lentsch, Caesar, and Gavrila]{lentsch2024union}
Ted Lentsch, Holger Caesar, and Dariu Gavrila.
\newblock {UNION}: Unsupervised 3d object detection using object appearance-based pseudo-classes.
\newblock In \emph{The Thirty-eighth Annual Conference on Neural Information Processing Systems}, 2024.

\bibitem[Li et~al.(2022)Li, Zhang, Lin, Wang, and Shen]{li2022rigidflow}
Ruibo Li, Chi Zhang, Guosheng Lin, Zhe Wang, and Chunhua Shen.
\newblock Rigidflow: Self-supervised scene flow learning on point clouds by local rigidity prior.
\newblock In \emph{Proceedings of the IEEE/CVF Conference on Computer Vision and Pattern Recognition}, pages 16959--16968, 2022.

\bibitem[Li et~al.(2023)Li, Zheng, Ferroni, Pontes, and Lucey]{li2023fast}
Xueqian Li, Jianqiao Zheng, Francesco Ferroni, Jhony~Kaesemodel Pontes, and Simon Lucey.
\newblock Fast neural scene flow.
\newblock In \emph{Proceedings of the IEEE/CVF International Conference on Computer Vision}, pages 9878--9890, 2023.

\bibitem[Lin and Caesar(2024)]{lin2024icp}
Yancong Lin and Holger Caesar.
\newblock Icp-flow: Lidar scene flow estimation with icp.
\newblock In \emph{Proceedings of the IEEE/CVF Conference on Computer Vision and Pattern Recognition}, pages 15501--15511, 2024.

\bibitem[Moosmann and Fraichard(2010)]{moosmann2010motion}
Frank Moosmann and Thierry Fraichard.
\newblock Motion estimation from range images in dynamic outdoor scenes.
\newblock In \emph{2010 IEEE International Conference on Robotics and Automation}, pages 142--147. IEEE, 2010.

\bibitem[Puy et~al.(2020)Puy, Boulch, and Marlet]{puy2020flot}
Gilles Puy, Alexandre Boulch, and Renaud Marlet.
\newblock Flot: Scene flow on point clouds guided by optimal transport.
\newblock In \emph{European conference on computer vision}, pages 527--544. Springer, 2020.

\bibitem[Qi et~al.(2017)Qi, Yi, Su, and Guibas]{qi2017pointnet++}
Charles~Ruizhongtai Qi, Li Yi, Hao Su, and Leonidas~J Guibas.
\newblock Pointnet++: Deep hierarchical feature learning on point sets in a metric space.
\newblock \emph{Advances in neural information processing systems}, 30, 2017.

\bibitem[Shi et~al.(2019)Shi, Wang, and Li]{Shi_2019_CVPR}
Shaoshuai Shi, Xiaogang Wang, and Hongsheng Li.
\newblock Pointrcnn: 3d object proposal generation and detection from point cloud.
\newblock In \emph{The IEEE Conference on Computer Vision and Pattern Recognition (CVPR)}, 2019.

\bibitem[Song and Yang(2022)]{NEURIPS2022_c6e38569}
Ziyang Song and Bo Yang.
\newblock Ogc: Unsupervised 3d object segmentation from rigid dynamics of point clouds.
\newblock In \emph{Advances in Neural Information Processing Systems}, pages 30798--30812. Curran Associates, Inc., 2022.

\bibitem[Sun et~al.(2018)Sun, Yang, Liu, and Kautz]{Sun2018PWC-Net}
Deqing Sun, Xiaodong Yang, Ming-Yu Liu, and Jan Kautz.
\newblock {PWC-Net}: {CNNs} for optical flow using pyramid, warping, and cost volume.
\newblock In \emph{CVPR}, 2018.

\bibitem[Sun et~al.(2020)Sun, Kretzschmar, Dotiwalla, Chouard, Patnaik, Tsui, Guo, Zhou, Chai, Caine, et~al.]{sun2020scalability}
Pei Sun, Henrik Kretzschmar, Xerxes Dotiwalla, Aurelien Chouard, Vijaysai Patnaik, Paul Tsui, James Guo, Yin Zhou, Yuning Chai, Benjamin Caine, et~al.
\newblock Scalability in perception for autonomous driving: Waymo open dataset.
\newblock In \emph{Proceedings of the IEEE/CVF conference on computer vision and pattern recognition}, pages 2446--2454, 2020.

\bibitem[Thomas et~al.(2021)Thomas, Agro, Gridseth, Zhang, and Barfoot]{thomas2021self}
Hugues Thomas, Ben Agro, Mona Gridseth, Jian Zhang, and Timothy~D Barfoot.
\newblock Self-supervised learning of lidar segmentation for autonomous indoor navigation.
\newblock In \emph{2021 IEEE International Conference on Robotics and Automation (ICRA)}, pages 14047--14053. IEEE, 2021.

\bibitem[Vedder et~al.(2023)Vedder, Peri, Chodosh, Khatri, Eaton, Jayaraman, Liu, Ramanan, and Hays]{vedder2023zeroflow}
Kyle Vedder, Neehar Peri, Nathaniel Chodosh, Ishan Khatri, Eric Eaton, Dinesh Jayaraman, Yang Liu, Deva Ramanan, and James Hays.
\newblock Zeroflow: Scalable scene flow via distillation, 2023.

\bibitem[Wang et~al.(2025)Wang, Feng, Jiang, Liu, and Wang]{10906337}
Guangming Wang, Zhiheng Feng, Chaokang Jiang, Jiuming Liu, and Hesheng Wang.
\newblock Unsupervised learning of 3d scene flow with lidar odometry assistance.
\newblock \emph{IEEE Transactions on Intelligent Transportation Systems}, pages 1--11, 2025.

\bibitem[Wang et~al.(2023)Wang, Gao, Han, Chen, Zhao, Lyu, and Hao]{wang2023active}
Shuaijun Wang, Rui Gao, Ruihua Han, Jianjun Chen, Zirui Zhao, Zhijun Lyu, and Qi Hao.
\newblock Active scene flow estimation for autonomous driving via real-time scene prediction and optimal decision.
\newblock \emph{IEEE Transactions on Intelligent Transportation Systems}, 2023.

\bibitem[Ward~Jr(1963)]{ward1963hierarchical}
Joe~H Ward~Jr.
\newblock Hierarchical grouping to optimize an objective function.
\newblock \emph{Journal of the American statistical association}, 58\penalty0 (301):\penalty0 236--244, 1963.

\bibitem[Wei et~al.(2021)Wei, Wang, Rao, Lu, and Zhou]{wei2021pv}
Yi Wei, Ziyi Wang, Yongming Rao, Jiwen Lu, and Jie Zhou.
\newblock Pv-raft: Point-voxel correlation fields for scene flow estimation of point clouds.
\newblock In \emph{Proceedings of the IEEE/CVF conference on computer vision and pattern recognition}, pages 6954--6963, 2021.

\bibitem[Wilson et~al.(2023)Wilson, Qi, Agarwal, Lambert, Singh, Khandelwal, Pan, Kumar, Hartnett, Pontes, et~al.]{wilson2023argoverse}
Benjamin Wilson, William Qi, Tanmay Agarwal, John Lambert, Jagjeet Singh, Siddhesh Khandelwal, Bowen Pan, Ratnesh Kumar, Andrew Hartnett, Jhony~Kaesemodel Pontes, et~al.
\newblock Argoverse 2: Next generation datasets for self-driving perception and forecasting.
\newblock \emph{arXiv preprint arXiv:2301.00493}, 2023.

\bibitem[Wu et~al.(2020)Wu, Wang, Li, Liu, and Fuxin]{wu2020pointpwc}
Wenxuan Wu, Zhi~Yuan Wang, Zhuwen Li, Wei Liu, and Li Fuxin.
\newblock Pointpwc-net: Cost volume on point clouds for (self-) supervised scene flow estimation.
\newblock In \emph{Computer Vision--ECCV 2020: 16th European Conference, Glasgow, UK, August 23--28, 2020, Proceedings, Part V 16}, pages 88--107. Springer, 2020.

\bibitem[Yi et~al.(2018)Yi, Huang, Liu, Kalogerakis, Su, and Guibas]{yi2018deep}
Li Yi, Haibin Huang, Difan Liu, Evangelos Kalogerakis, Hao Su, and Leonidas Guibas.
\newblock Deep part induction from articulated object pairs.
\newblock \emph{arXiv preprint arXiv:1809.07417}, 2018.

\bibitem[Zhai et~al.(2021)Zhai, Xiang, Lv, and Kong]{zhai2021optical}
Mingliang Zhai, Xuezhi Xiang, Ning Lv, and Xiangdong Kong.
\newblock Optical flow and scene flow estimation: A survey.
\newblock \emph{Pattern Recognition}, 114:\penalty0 107861, 2021.

\bibitem[Zhai et~al.(2025)Zhai, Bao, and Xiang]{zhai2025dmrflow}
Mingliang Zhai, Bing-Kun Bao, and Xuezhi Xiang.
\newblock Dmrflow: 4d radar scene flow estimation with decoupled matching and refinement.
\newblock \emph{IEEE Transactions on Circuits and Systems for Video Technology}, 2025.

\bibitem[Zhang et~al.(2024{\natexlab{a}})Zhang, Yang, Fang, Geng, and Jensfelt]{zhang2024deflow}
Qingwen Zhang, Yi Yang, Heng Fang, Ruoyu Geng, and Patric Jensfelt.
\newblock Deflow: Decoder of scene flow network in autonomous driving.
\newblock \emph{arXiv preprint arXiv:2401.16122}, 2024{\natexlab{a}}.

\bibitem[Zhang et~al.(2024{\natexlab{b}})Zhang, Yang, Li, Andersson, and Jensfelt]{zhang2024seflow}
Qingwen Zhang, Yi Yang, Peizheng Li, Olov Andersson, and Patric Jensfelt.
\newblock Seflow: A self-supervised scene flow method in autonomous driving.
\newblock In \emph{European Conference on Computer Vision}, pages 353--369. Springer, 2024{\natexlab{b}}.

\end{thebibliography}
}

\end{document}